\documentclass[letterpaper]{article} 
\usepackage{aaai25}  
\usepackage{times}  
\usepackage{helvet}  
\usepackage{courier}  
\usepackage[hyphens]{url}  
\usepackage{graphicx} 
\usepackage{natbib}  
\usepackage{caption} 
\usepackage{algorithm}
\usepackage{algorithmic}
\usepackage{booktabs}
\usepackage{subfigure,color}
\usepackage{multirow,amsmath}
\usepackage{array}
\usepackage{amssymb}
\usepackage{multirow,amsmath}
\usepackage{tabulary}
\usepackage{pgfplots}

\usepackage{newfloat}
\usepackage{listings}
\usepackage{dsfont}

\usepackage{newfloat}
\usepackage{listings}
\DeclareCaptionStyle{ruled}{labelfont=normalfont,labelsep=colon,strut=off} 
\floatstyle{ruled}
\newfloat{listing}{tb}{lst}{}
\floatname{listing}{Listing}
\title{Explicit Relational Reasoning Network for Scene Text Detection}

\author {
    Yuchen Su\textsuperscript{\rm 1},
    Zhineng Chen\textsuperscript{\rm 1},
    Yongkun Du\textsuperscript{\rm 1},
    Zhilong Ji\textsuperscript{\rm 2},
    Kai Hu\textsuperscript{\rm 3},
    Jinfeng Bai\textsuperscript{\rm 2},
    Xieping Gao\textsuperscript{\rm 4 \dag}
}
\affiliations {
    \textsuperscript{\rm 1} School of Computer Science, Fudan University \\ 
    \textsuperscript{\rm 2} Tomorrow Advancing Life \\
    \textsuperscript{\rm 3} School of Computer Science, Xiangtan University\\
    \textsuperscript{\rm 4} Laboratory for Artificial Intelligence and International Communication, Hunan Normal University \\
    \{ycsu23, ykdu23\}@m.fudan.edu.cn,
    zhinchen@fudan.edu.cn, zhilongji@hotmail.com, jfbai.bit@gmail.com,  kaihu@xtu.edu.cn,  xpgao@hunnu.edu.cn
}

\usepackage{bibentry}

\begin{document}

\maketitle
\let\thefootnote\relax\footnotetext{$^{\dag}$ Corresponding author.}

\begin{abstract}

Connected component (CC) is a proper text shape representation that aligns with human reading intuition. However, CC-based text detection methods have recently faced a developmental bottleneck that their time-consuming post-processing is difficult to eliminate. To address this issue, we introduce an explicit relational reasoning network (ERRNet) to elegantly model the component relationships without post-processing. Concretely, we first represent each text instance as multiple ordered text components, and then treat these components as objects in sequential movement. In this way, scene text detection can be innovatively viewed as a tracking problem. From this perspective, we design an end-to-end tracking decoder to achieve a CC-based method dispensing with post-processing entirely. Additionally, we observe that there is an inconsistency between classification confidence and localization quality, so we propose a Polygon Monte-Carlo method to quickly and accurately evaluate the localization quality. Based on this, we introduce a position-supervised classification loss to guide the task-aligned learning of ERRNet. Experiments on challenging benchmarks demonstrate the effectiveness of our ERRNet. It consistently achieves state-of-the-art accuracy while holding highly competitive inference speed.

\end{abstract}

\section{Introduction}

Scene text detection aims to locate text regions within images. It is a fundamental step for many computer vision and artificial intelligence tasks \cite{zhang2021character,wei2022textblock,fang2022abinet++,wang2022omnivl,meng2022adavit,zhang2024choose}.
Despite recent advancements, detecting text in the wild remains challenging due to the varied scales, shapes, colors, and fonts of text.

\begin{figure}[t]
    \centering
    \subfigure [\scriptsize TextBPN++ \cite{zhang2023arbitrary}]
    {
        \includegraphics[width=0.22\textwidth,height=2.0cm]{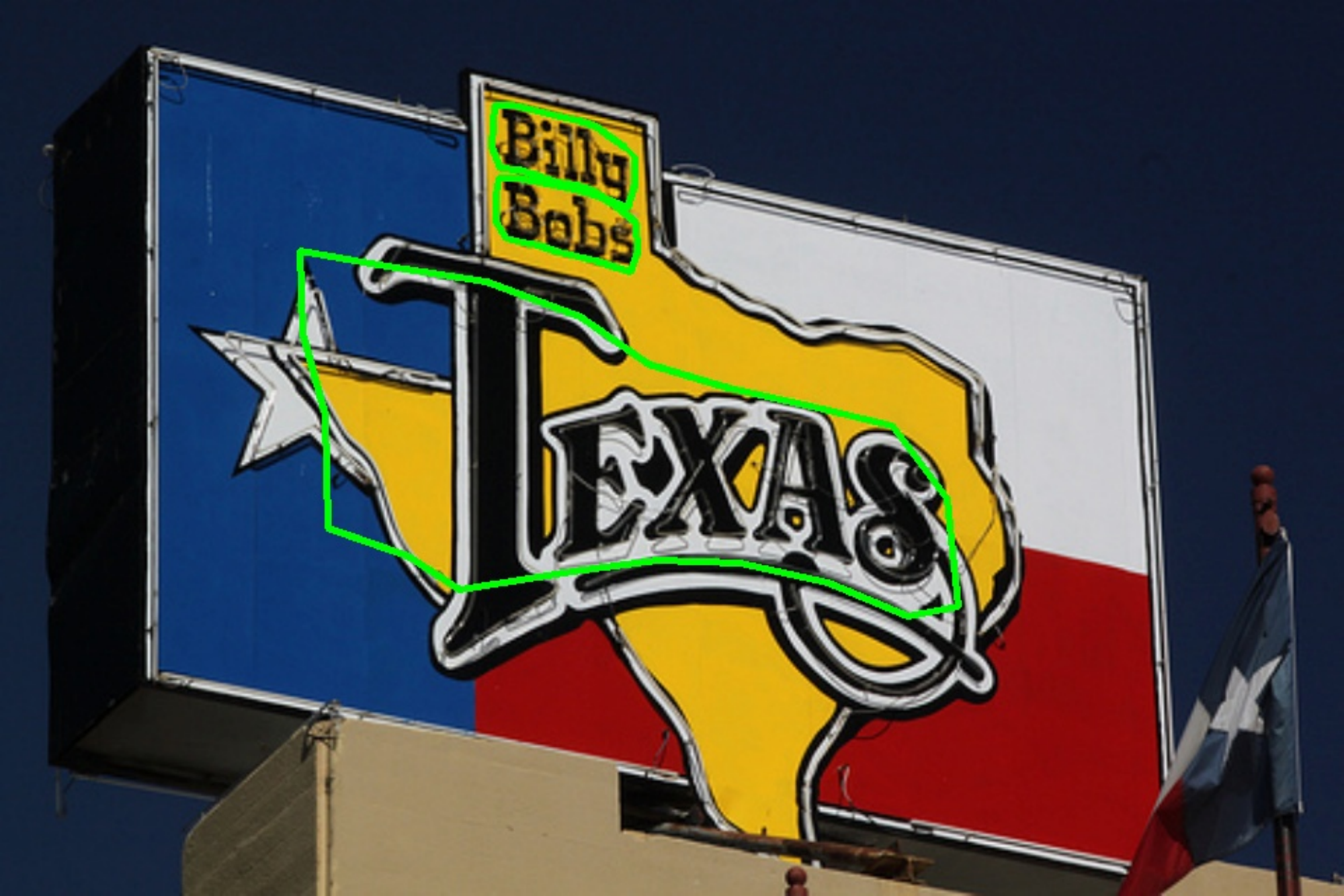}
        \label{fig-TextRay}
    }
    \subfigure [ \scriptsize LRANet \cite{su2024lranet}]
    {
        \includegraphics[width=0.22\textwidth,height=2.0cm]{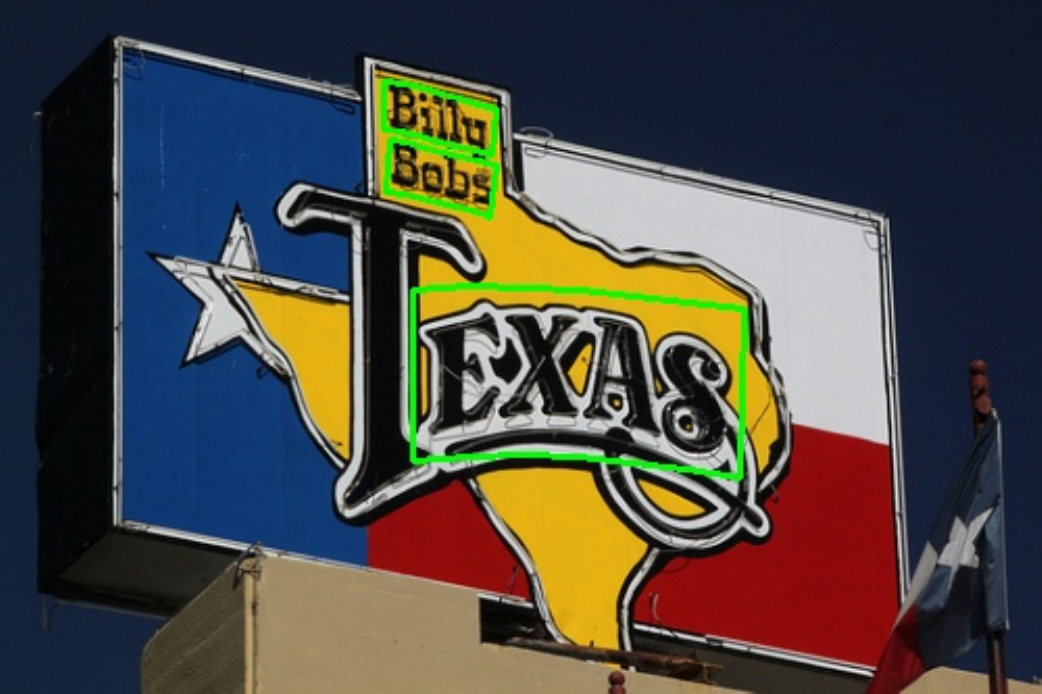}
        \label{fig-TextDCT}
    } 
    \vspace{-2mm}
    
    \subfigure [\scriptsize DRRG \cite{2020Deep}]
    {
        \includegraphics[width=0.22\textwidth,height=2.0cm]{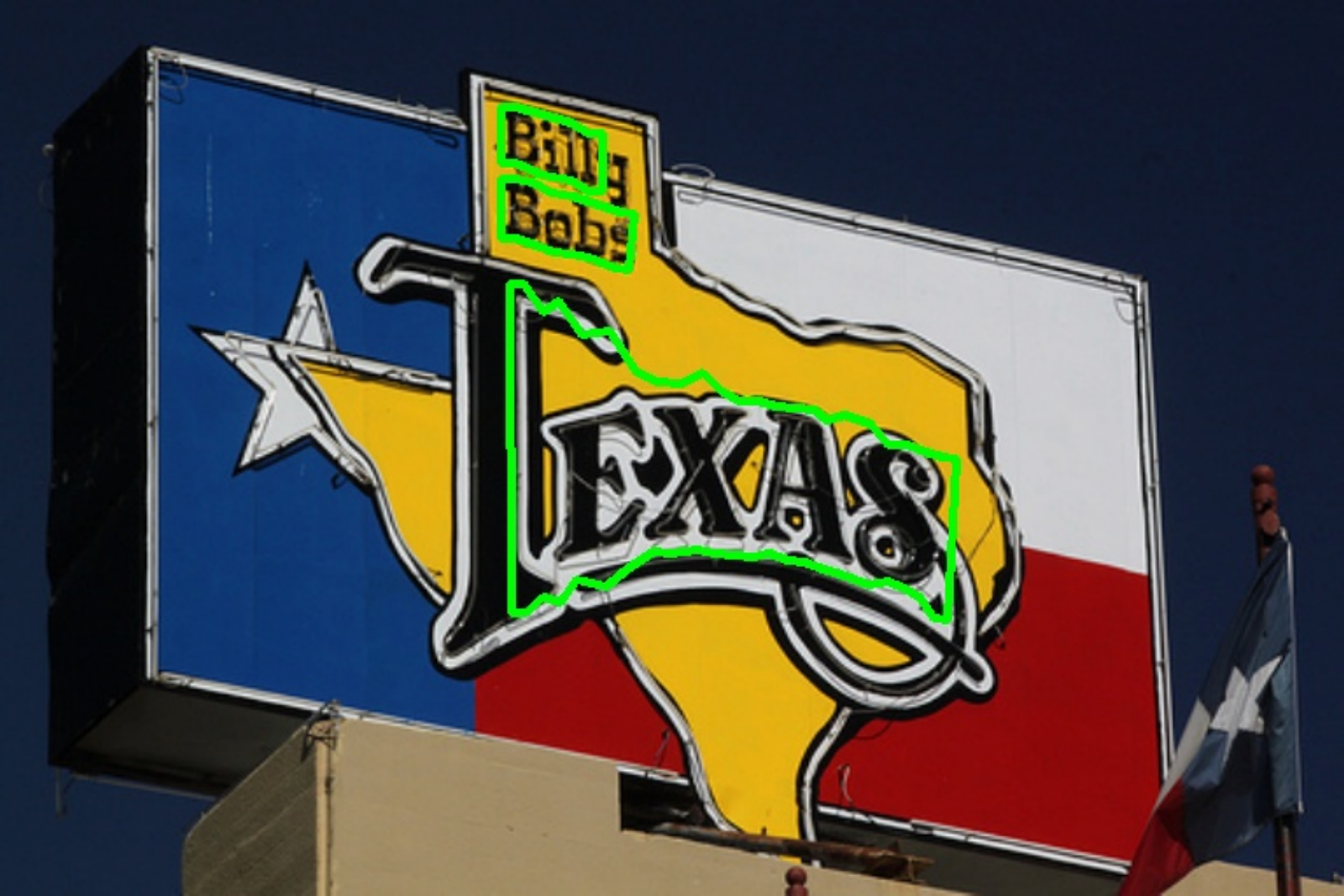}
        \label{fig-FCENet}
    }
    \subfigure [\scriptsize \textbf{Ours}]
    {
        \includegraphics[width=0.22\textwidth,height=2.0cm]{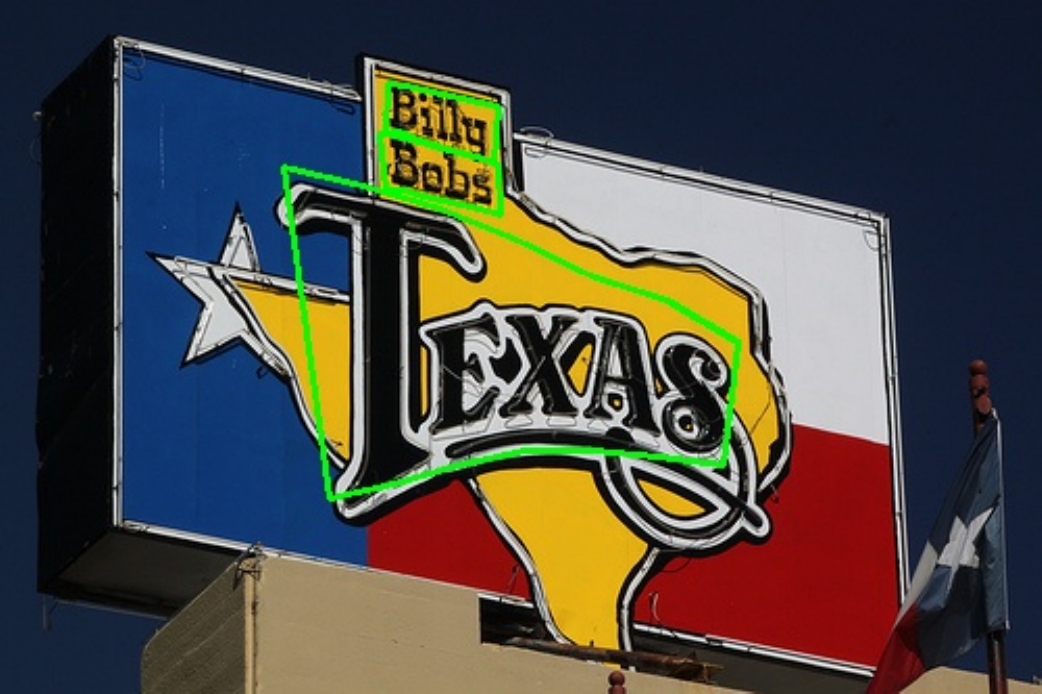}
        \label{fig-lranet}
    }
     \vspace{-3mm}
     
    \caption{Comparison with leading text detection methods of three different types: (a) Segmentation-based method (TextBPN++), (b) Regression-based method (LRANet), and (c) Connected component-based method (DRRG).
    }
    \label{fig:compare1}
    \vspace{-2mm}
    
\end{figure}

Segmentation- and regression-based methods are two mainstream scene text detection methods. The former \cite{liao2022real,zhao2024cbnet} utilize shrunk text kernel to separate adhesive text instances and cluster text pixels into distinct instances through heuristic post-processing. Although this pixel-level representation approach can flexibly fit arbitrary-shaped text, it overly focuses on local textual cues, leading to sensitivity to local noise, as illustrated in Fig.~\ref{fig:compare1}(a). In contrast, regression-based methods \cite{wang2022tpsnet,su2024lranet} regress parameterized text shapes to directly capture the text's overall geometric layout, which have higher resistance to local noise. However, these methods lack scale and shape invariance, as text scale and shape show great variability, making it difficult to directly perceive the overall geometric layout of complex text with accuracy, as shown in Fig.~\ref{fig:compare1}(b). 

\begin{figure}[t]
    \centering
\includegraphics[width=\linewidth]{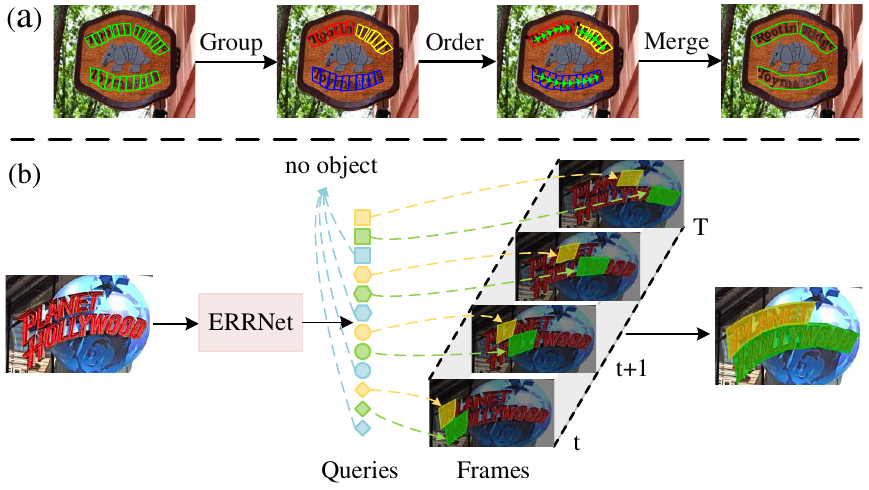}
  \caption{(a) Post-processing illustration of typical CC-based methods, which require grouping and ordering components one by one. (b) The pipeline of ERRNet, which has no post-processing. ERRNet views each text instance as multiple text components in sequential movement. Same shape queries indicate predictions in different text instances but in the same sequential position, same color queries represent predictions in one instance, and temporal relationships denote the sequential relationships of components.}

  \label{fig:overview}
      \vspace{-2mm}
\end{figure}

From a hybrid perspective, connected component (CC)-based methods \cite{2020Deep}, which treat each text instance as a combination of a series of adjacent text components, are a reasonable integration of the above two types of methods. Compared to complex text contours, text components have fixed shapes (\textit{e.g.}, circles \cite{long2018textsnake}, quadrilaterals \cite{feng2019textdragon}), and smaller size variations. Meanwhile, components are more resistant to local noise than individual pixels. However, CC-based methods have been less studied recently due to their tricky and time-consuming post-processing, which includes grouping components one by one based on their associative relationships to differentiate between text instances, and then ordering internal components within each instance individually according to their sequential relationships, as depicted in Fig.~\ref{fig:overview}(a). 

To address this issue, we formulate scene text detection as a tracking problem for the first time. As shown in Fig.~\ref{fig:overview}(b), each text instance is decomposed into multiple ordered text components. The instances are then represented by a series of motion frames, each containing a specified text component of every text instance. In each frame, we predict components across different text instances in a consistent sequence, where temporal relationships mirror the sequential relationships of components. By leveraging this conceptualization, we can transform the CC-based detection into an end-to-end tracking task, dispensing with post-processing entirely.

To this end, we introduce an explicit relational reasoning (ERR) decoder that seamlessly integrates relational prediction of components within the framework of positional regression. As shown in Fig.~\ref{fig:overview}(b), the same shape queries represent predictions in different text instances but in the same frame; the same color queries indicate predictions belonging to the same instance; and each frame has the number of predictions for parallel processing. To achieve this goal, we employ bipartite graph matching between the output component sequence and the ground-truth component sequence for each text instance, and supervise the sequence as a whole.

Additionally, we observe that text detection methods encounter a misalignment issue between classification and localization tasks, resulting in either high classification confidence with relatively low localization quality or vice versa. Existing studies mostly overlook this issue, mainly because efficiently evaluating the localization quality of arbitrary-shaped text detection results is difficult. To address this, we propose a Polygon Monte-Carlo method to quickly and accurately calculate the Polygon Intersection over Union (PIoU) between predicted results and ground-truth. Based on this, we introduce a position-supervised classification loss to better guide the task-aligned learning.

Building upon these designs, we propose an explicit relational reasoning network, termed ERRNet. It first generates initial text components via a text component initialization module, and then directly outputs the position of each component and the relationship among different components in order, achieving accurate and efficient scene text detection. The main contributions of this paper are as follows:

    $\bullet$ We propose ERRNet, a much simpler and faster CC-based text detection method. It eliminates the complex post-processing by innovatively modeling the component relationships from a tracking perspective.

    $\bullet$ We introduce a position-supervised classification loss to force the classification confidence and localization quality of text instances to be consistent, guiding the detector better trained and thereby enhancing the detection performance.
    
    $\bullet$ Extensive experiments are conducted on challenging benchmarks, which demonstrate that ERRNet is the most accurate detector and it also ranks among the fastest detectors.

\begin{figure*}
\centering
\includegraphics[width=\textwidth]{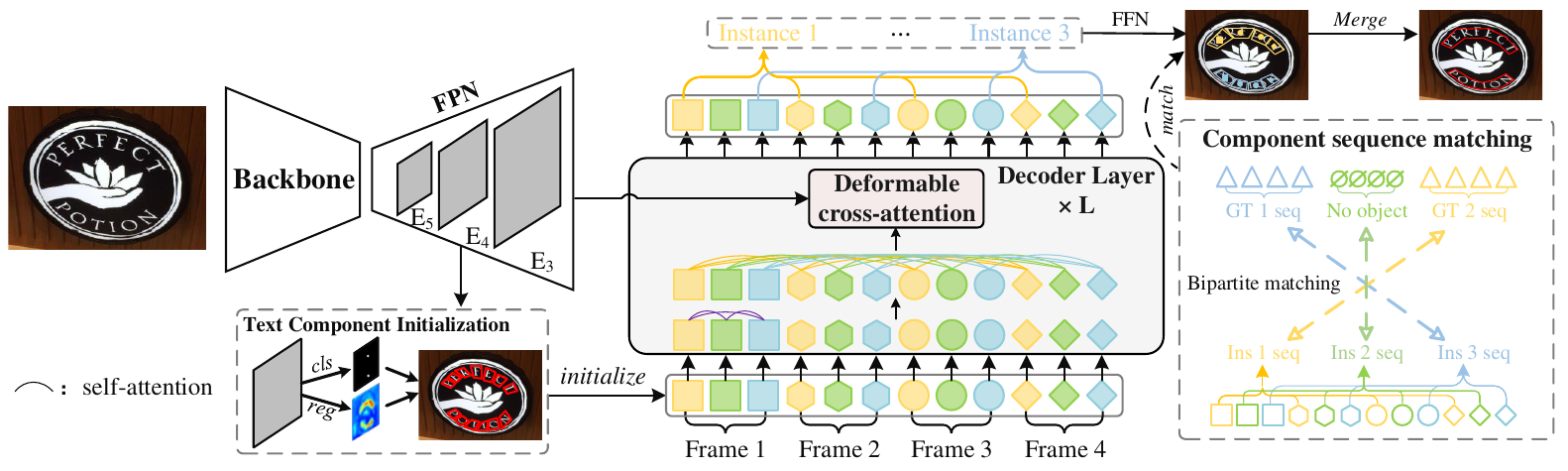}
\caption{The architecture of ERRNet, which is mainly composed of three modules: (a) the backbone and feature pyramid network (FPN) for multi-scale feature extraction, (b) the text component initialization module to generate initial component queries, and (c) the explicit relational reasoning decoder for decoding the component sequence for each text instance in order.}
\label{fig:architecture}
\end{figure*}

\section{Related Work}

\subsection {Segmentation-Based Methods}
Segmentation-based methods \cite{xu2019textfield,zhang2021adaptive,yang2023text} view text detection as an image segmentation problem, which usually adopt text kernels to separate adhesive text instances and expand them with heuristic post-processing. For example, 
DB \cite{liao2020Real} and its improved version DB++ \cite{liao2022real} introduce a differentiable binarization module that assigns higher thresholds to text boundaries, reinforcing the distinction between adjacent text instances. CBNet \cite{zhao2024cbnet}
proposes a context-aware module to 
enhance the text kernel segmentation results and a boundary-guided module to expand the text kernel in a learnable manner. Although these pixel-level modeling methods can flexibly fit arbitrary-shaped text, they usually need computationally intensive post-processing to reconstruct text boundaries, and are sensitive to text-like backgrounds due to neglecting the geometric context of holistic text instances.

\subsection {Regression-Based Methods}
Regression-based methods \cite{liu2020abcnet,he2021most,su2024lranet} treat scene text detection as a special type of object detection. Earlier methods \cite{liao2017textboxes,zhou2017east,lyu2018multi} use modified anchor-mechanisms to detect multi-oriented text instances. For example, Textboxes \cite{liao2017textboxes} increases the proportion of anchor-boxes to adapt to varied text scales.

To detect irregularly shaped text, some parameterized text shape methods are proposed. For example, ABCNet \cite{liu2020abcnet} utilizes Bernstein polynomial to convert the long sides of text into Bezier curves. 
LRANet \cite{su2024lranet}
leverages a linear combination of pre-defined eigenvectors to represent text boundaries. However, only optimizing the regression target is insufficient, as text scale and shape have great variability, accurately perceiving the overall text layout requires a meticulously designed network that provides a large receptive field. In response, CT-Net \cite{shao2023ct} proposes multi-stage contour refinement modules to iteratively and adaptively refine text contours. Similarly, DPText-DETR \cite{ye2023dptext} employs a Transformer framework to address complex text layouts by capturing long-range contextual dependencies. However, the complex structure of these methods hinders the further development of efficient and accurate text detectors following this pipeline.

\subsection {CC-Based Methods}
Connected component (CC)-based methods \cite{tian2016detecting,long2018textsnake,2020Deep} can be viewed as a middle ground between segmentation-based and regression-based methods, where the basic shape representation unit is a text part or character, followed by a linking-based post-processing procedure for generating final text boundaries. Before the era of deep learning, CC-based methods \cite{yin2013robust,sun2015robust} had been widely used in scene text detection, as CC is an ideal text shape representation that aligns with human reading intuition. In recent years, CTPN \cite{tian2016detecting} 
uses horizontal text components with a fixed-size width for handling text instances with extreme aspect ratios. TextSnake \cite{long2018textsnake}  
introduces an ordered set of disks to represent text instances, and adopts a segment network to learn the relationship among disks.
DRRG \cite{2020Deep} introduces a graph convolutional network (GCN) to learn the associative relationships of text components for grouping, and uses the shortest path algorithm to infer the sequential relationship of components in each group. ReLaText \cite{ma2021relatext} formulates CC-based methods as a scene graph generation task, and utilizes GCN to learn the relationships among pre-defined triplets.

Although CC is a better text shape representation due to its stability in size and shape, and its flexibility in representing text of arbitrary shapes, the complex and tricky post-processing has slowed the progress of CC-based methods. Therefore, we formulate the CC-based detection into an end-to-end tracking-like pipeline to eliminate post-processing, aiming to make CC-based methods shine again.

\section{Methodology}

\subsection{Overview}

The overall structure of ERRNet is illustrated in Fig.~\ref{fig:architecture}. Given an image with text, ERRNet first employs a ResNet-50 \cite{he2016deep} with DCN \cite{zhu2019deformable} as the backbone network, followed by a feature pyramid network (FPN) to extract multi-scale feature maps. Subsequently, a text component initialization (TCI) module generates initial coarse text components. These components are then sent into a Transformer decoder, which further refines them and establishes their associative and sequential relationships. Finally, the text components are aggregated into holistic text instances according to the explicit reasoning results. Besides, we also discuss how to better align the classification and localization tasks from the loss perspective and use it to train ERRNet better.

\subsection {Text Component Initialization}

\begin{figure}[t]
    \centering
    \includegraphics[width=\linewidth]{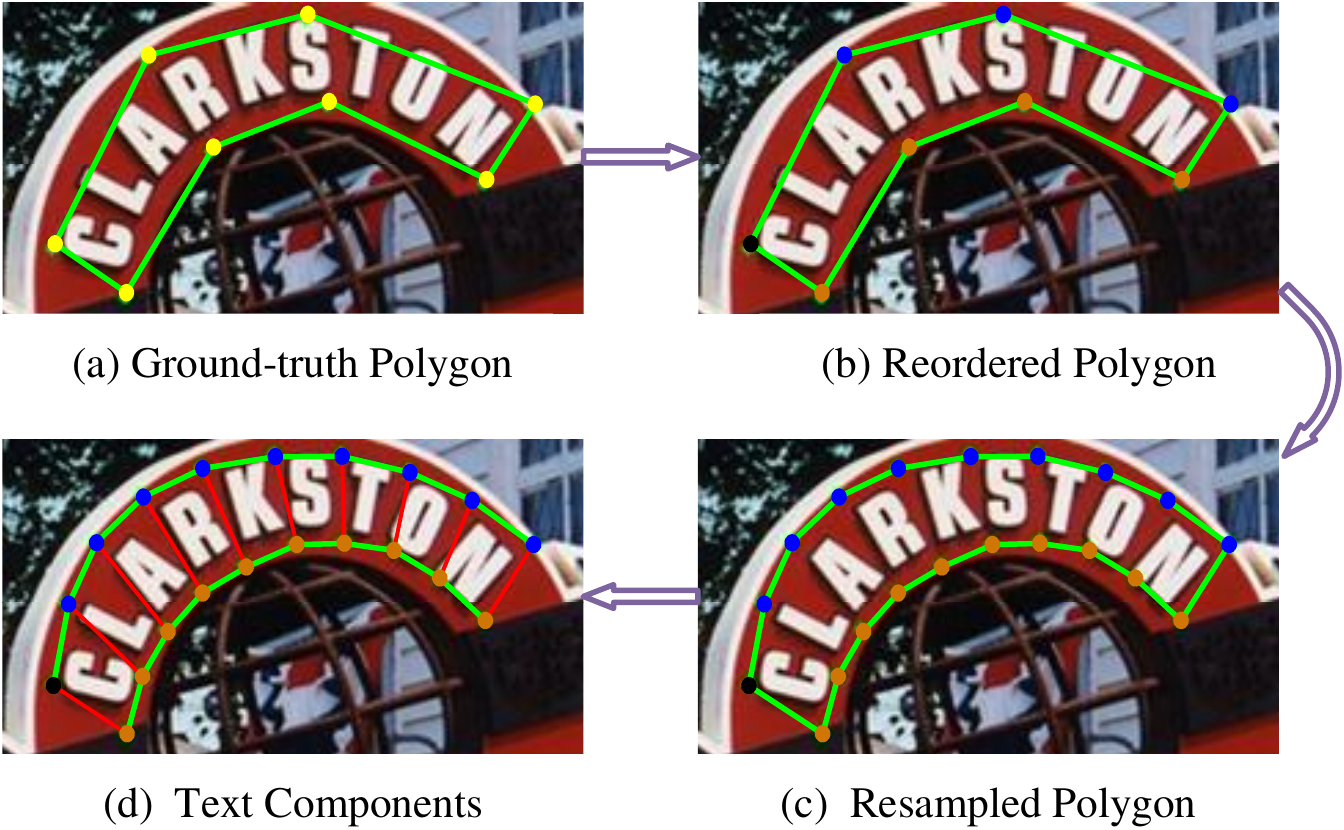}
    \caption{Illustration of ground-truth text component generation for a text primitive. Black point means start point.} 

    \label{fig:component}
\end{figure}

In our work, each text boundary is represented using a series of ordered quadrilateral components, and each text component is represented by four vertices, as shown in Fig.~\ref{fig:component}.
To delineate components within each text instance, we first apply the method in \cite{wang2022tpsnet} to divide the text contour into two long sides and determine the order and starting point of the contour points, as shown in Fig.~\ref{fig:component}(b). Subsequently, we sample $m$ points on each long side to divide each text instance into a series of components ordered along these long sides. Here, we adopt B-spline interpolation for point sampling because its strong local control capability allows for better fitting of complex text shapes. To elaborate, we construct a B-spline curve with the following formula:
\begin{equation}
C(u)=\sum_{i=0}^{\bar{n}} N_{i, k}(u) \bar{P}_i \,,
\end{equation}
where $\bar{P}_i$ is the $i$-$th$ vertex in ground-truth long side, $N_{i, k}(u)$ is the B-spline curve basis function of degree $k$:
\begin{equation}
N_{i, 0}(u)= \begin{cases}1 & \text { if } u_i \leq u<u_{i+1} \\ 0 & \text { otherwise }\end{cases} \,,
\end{equation}
\begin{equation}
\begin{split}
N_{i, k}(u) = & \frac{u - u_i}{u_{i+k} - u_i} N_{i, k-1}(u) \\
& + \frac{u_{i+k+1} - u}{u_{i+k+1} - u_{i+1}} N_{i+1, k-1}(u)
\end{split}
\end{equation}
where $k=3$, and $u_i$ denotes the $i$-$th$ knot, with the half-open interval $\left[u_i, u_{i+1}\right)$
representing the $i$-$th$ knot span. Next, we equally sample $m$ vertices on the B-spline curve (see Fig.~\ref{fig:component}(c)):
\begin{equation}
P_j=C\left(u_j\right)=\sum_{i=1}^{\bar{n}} \bar{P}_i N_{i, k}\left(u_j\right), \quad j=1,2, \ldots, m \,.
\end{equation}
Thus, each text instance can be divided into a series of ordered quadrilateral regions based on the sampling points and pre-defined directions, as shown in Fig.~\ref{fig:component}(d).

Following \cite{su2024lranet}, we adopt a lightweight module to predict the text components. Specifically, after each output layer of the FPN, two sets of $3 \times 3$ convolution layers are utilized to extract classification features $F_{cls}$ and regression features $F_{reg}$, respectively. Then, distinct $3 \times 3$ convolutions are applied to $F_{cls}$ and $F_{reg}$ for classifying and regressing the interpolated and sorted text contours. Finally, we organize these text contours into a sequence of ordered quadrilateral components. Each component $F_{comp}\in{\mathbb{R}^{{1 \times 4c_v}}}$ is formed by concatenating the positional encodings of its four vertices along the channel dimension, where $c_v$ denotes the number of channels per vertex.

\subsection {Explicit Relational Reasoning Decoder}

CC-based methods need grouping and ordering text components to construct the text boundaries, as shown in Fig.~\ref{fig:overview}(a). Previous methods \cite{long2018textsnake,2020Deep} handle these components sequentially in a non-parallel fashion. To get rid of this computationally intensive post-processing, we develop a unique perspective of treating scene text detection as a tracking problem, where scene text is conceptualized as text components in sequential movement. From this viewpoint, we propose an explicit relational reasoning strategy to reformulate CC-based methods as an end-to-end tracking task and eliminate post-processing.

Specifically, we first select top-$n$ groups of components from the TCI module based on descending classification scores. Notably, $n$ is typically larger than the number of instances in any given image. Next, these components are organized into a series of hypothetical frames $F =(F_1^1,F_1^2, \cdots, F_1^n),\cdots,(F_t^1,F_t^2, \cdots, F_t^n)$, where $t$ and $n$ denote the total number of frames and the number of components in each frame, respectively. 
The frames satisfy: 1) $F_i^1, F_i^2,\cdots, F_i^n$ have the same ordinal position but in different text instances; 2) $F_1^j, F_2^j,\cdots, F_t^j$ belong to the same instance, maintaining an internal order consistent with their temporal sequence from 1 to $t$. Through this definition, we effectively map the associative and sequential relationships among text components onto  positional relations within each frame and temporal relations across these frames.

When the output order aligns with the above pre-defined order, we can directly distinguish text instances based on the content of each frame and determine the order of components within each text instance according to the temporal relationships, thereby directly getting the text predictions. To achieve this, we introduce a component sequence matching that supervises the component sequence as a whole.

\noindent \textbf{Component Sequence Matching.}
As the ERRNet decodes $n$ components each frame, the number of text instances formed by component sequence is also $n$. We denote the predicted components sequences as $\hat{y} = \{\hat{y}_i\}_{i=1}^{n}$ and define the ground-truth set of component sequence as $y$, which consists of $n$ elements padded with $\emptyset$. To find an optimal pair-wise matching between these two sets, we seek a permutation of $n$ elements that minimizes the cost:
\begin{equation}
\hat{\sigma}={\arg \min } \sum_i^n \mathcal{L}_{\text {match }}\left( \hat{y}_{\sigma(i)}, y_i\right) \,,
\end{equation}
where $\mathcal{L}_{\text {match}}\left( \hat{y}_{\sigma(i)}, y_i\right)$ represents the matching cost between the predicted component sequence indexed by $\sigma(i)$ and
the ground-truth $y_i$.

Given the $N = n \cdot t$ quadrilateral predictions for the component prediction sequence, we can associate $n$ component sequences for each instance  based on their location indices, as illustrated by $Ins 1 seq \dots Ins 3 seq$ in Fig.~\ref{fig:architecture}. 
The ground-truth for the $i$-$th$ instance can be represented as follows:
\begin{equation}
y_i=\left\{\left(c_i, c_i \ldots, c_i\right),\left(q_{i, 0}, q_{i, 1} \ldots, q_{i, t}\right)\right\} \,,
\end{equation}
where $c_i$ is the target class label ($0$ for text and $1$ for $\emptyset$), and $q_{i, t}$ is a vector that specifies the ground-truth of quadrilateral component locations. For the predictions of component sequence with index $\sigma(i)$, we denote the predicted classification scores as:
\begin{equation}
\hat{p}_{(\sigma(i))}\left(c_i\right) = \left\{\hat{p}_{(\sigma(i), 0)}\left(c_i\right) \ldots, \hat{p}_{(\sigma(i), t)}\left(c_i\right)\right\} \,,
\end{equation}
and the predicted component sequence positions as:
\begin{equation}
\hat{q}_{\sigma(i)}=\left\{\hat{q}_{(\sigma(i), 0)}, \hat{q}_{(\sigma(i), 1)} \ldots, \hat{q}_{(\sigma(i), t)}\right\} \,.
\end{equation}

Using the above notation, we define the matching loss as:
\begin{equation}
\mathcal{L}_{\text {match }}\left(y_i, \hat{y}_{\sigma(i)}\right)=\mathcal{L}_{\text {cls}}\left(c_i, \hat{p}_{\sigma(i)}\left(c_i\right)\right) + \mathcal{L}_{\text {reg}}\left(q_i, \hat{q}_{\sigma(i)}\right) \,,
\end{equation}
where $\mathcal{L}_{\text {cls}}$ and $\mathcal{L}_{\text {reg}}$ denote the Focal loss \cite{lin2017focal} and $\ell_1$ loss, respectively. Finally, we can find a one-to-one matching between the sequences using the Hungarian algorithm \cite{kuhn1955hungarian}, thereby maintaining the relative positions of predictions for the same instance across different frames.

\noindent \textbf{Decoder Structure.}
From a tracking perspective, we only need to model the relationships between objects in the first frame and the temporal relationships between different frames. Thus, we adopt a modulated transformer decoder module for parallel component sequence decoding. Specifically, we initially incorporate an intra-frame self-attention module to model the spatial position among component queries within the first frame. Subsequently, we use an inter-frame self-attention module to model the temporal relationship between distinct frames. The outputs are then fed into a multi-scale deformable cross-attention module \cite{zhu2020deformable} for interacting features with the flattened output layers of FPN. Finally, these features are individually projected into task-specific space using a feedforward network (FFN).  Notably, each point in  components allows channel-level interactions via FFN, as the initial components are formed by concatenating the positional encodings of their four vertices along the channel dimension.

\subsection{Task Alignment Learning}

Current text detection methods overlook the inconsistent prediction issue, i.e., a high classification score with a relatively low localization precision, and vice versa. A position-supervised classification loss could be a good solution. However, efficiently evaluating the localization quality of arbitrary-shaped text detection results is challenging. Thus, we propose a Polygon Monte-Carlo method to quickly and accurately calculate the Polygon Intersection over Union (PIoU) between predicted results and ground-truth.

\begin{figure}[t]
    \centering
    \includegraphics[width=\linewidth]{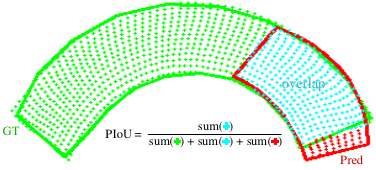}
    \caption{Illustration of our Polygon Monte-Carlo method for calculating the Polygon Intersection over Union (PIoU).} 
    \label{fig:polygon_iou}
\end{figure}

\noindent \textbf{Polygon Monte-Carlo Method.}
It comprises three steps. First, given a predicted text instance $\hat{G}$ and corresponding ground-truth $G$, we adopt TPS-align \cite{wang2022tpsnet} to sample $K$ points within the text instances. Next, we quantify these sampled points based on a pre-defined tolerance to disregard minor numerical discrepancies. Finally, we count the repeated elements among the sampled points in both the predicted instance $\hat{G}$ and the ground-truth $G$ as the intersection, and the union is calculated by adding these repeated points to the unique points from both instances, as shown in Fig.~\ref{fig:polygon_iou}. 
Notably, all the above steps are organized as matrix operations on the GPU. This enables us to set the sampling number $K$ to a large value, \textit{e.g.}, $10,000$, for an accurate approximation of the PIoU, and allows us to evaluate thousands of localization qualities shortly.

\noindent \textbf{Position-Supervised Classification Loss.}
Based on the calculated PIoU, we naturally use it to dynamically adjust classification targets for component queries, which smooths the training target and strengthens the correlation between high classification confidence and high-quality prediction. Thus, the position-supervised loss is expressed as:
\begin{equation}
\label{loss_qabce}
\begin{split}
\mathcal{L}_{\mathrm{cls}} = & \sum_{i=1}^{N_{\mathrm{pos}}} \left|s_i^\alpha - \hat{c}_i\right|^\gamma \mathrm{BCE}(\hat{c}_i, s_i^\alpha) \\
& \hspace{1em} + \sum_{j=1}^{N_{\mathrm{neg}}} \left|\hat{c}_j\right|^\gamma \mathrm{BCE}(\hat{c}_j, 0),
\end{split}
\end{equation}

\noindent where $\hat{c}$ is the predicted classification score, $s^i$ is the PIoU between the $i$-th ground-truth and its corresponding prediction, $\alpha$ is a scaling factor and $\gamma$ is a focusing parameter.

\noindent \textbf{Overall Loss.}
The full loss function is expressed as:
\begin{equation}
\label{loss_full}
\mathcal{L}=\mathcal{L}_{tci}+ \mathcal{L}_{dec}\,,  
\end{equation}
Where $\mathcal{L}_{tci}$ and $\mathcal{L}_{dec}$ refer to the losses for the TCI module and the ERR decoder, respectively. Both of these contain a classification loss $\mathcal{L}_{cls}$ and a regression loss $\mathcal{L}_{reg}$ for classifying and regressing text components. Here, the $\ell_1$ loss is applied to $\mathcal{L}_{reg}$ and $\mathcal{L}_{cls}$ denotes our proposed position-supervised classification loss.

\section{Experiments}

\begin{table*}[t]
  \centering
  \footnotesize
  \caption{Quantitative detection results on typical benchmarks. ``Seg'', ``Reg'' and ``CC'' means segmentation-, regression- and connected components-based methods. All listed FPS is measured from a
  single NVIDIA RTX3090 GPU.}
  \resizebox{0.95\linewidth}{!}{%
    \begin{tabular}{@{}l cc ccc ccc cccc  @{}}
      \toprule
      \multirow{2}*{Method} & \multirow{2}*{Type}  & \multirow{2}*{Ext} & \multicolumn{3}{c}{MSRA-TD500} & \multicolumn{3}{c}{Total-Text} &
      \multicolumn{4}{c}{CTW1500}  \\
      \cmidrule(r){4-6} \cmidrule(r){7-9} \cmidrule{10-13}
        & & & R  & P & F  & R  & P & F  & R  & P & F &FPS \\
      \midrule
    DB  \cite{liao2020Real}        & Seg   & $\checkmark$    & $79.2$         & $91.5$      & $84.9$     & $82.5$      & $87.1$       & $84.7$      & $80.2$     & $86.9$        & $83.4$ & $33.5$          \\
    TextBPN   \cite{zhang2021adaptive} & Seg   & $\checkmark$         & $84.5$                 & $86.6$      & $85.6$   & $85.2$      & $90.8$       & $87.9$     & $83.6$                 & $86.5$      & $85.0$  & $18.1$         \\
    FSG \cite{tang2022few}     & Seg   & $\checkmark$     & $84.8$                 & $91.6$      & $88.1$    & $85.7$      & $90.7$       & $88.1$     & $82.4$                 & $88.1$      & $85.2$   & --         \\  
    TextPMs \cite{zhang2022arbitrary}  & Seg   & $\checkmark$        & $87.0$                 & $91.0$      & $88.9$      & $87.7$      & $90.0$       & $88.8$      & $83.8$                 & $87.8$      & $85.7$    & $14.4$         \\ 
   TextBPN++ \cite{zhang2023arbitrary} & Seg   & $\checkmark$       & $86.8$                 & $93.7$      & $90.1$      & $\textbf{87.9}$      & $92.4$   & $\textbf{90.1}$      & $84.7$                 & $88.3$      & $86.5$    & $13.9$   \\
    CBNet \cite{zhao2024cbnet} & Seg  & $\checkmark$        & $84.8$                 & $91.1$      & $87.8$      & $82.5$      & $90.1$       & $86.1$      & $81.9$                 & $89.0$      & $85.3$    & --     
   \\ 
    
    \midrule
    ABCNet v2 \cite{liu2021abcnet}  & Reg & $\checkmark$             & $81.3$                 & $89.4$      & $85.2$   & $84.1$      & $89.2$       & $87.0$  &  $83.8$                 & $85.6$      & $84.7$  & --        \\
    TextDCT \cite{su2022textdct}   & Reg & --    & --                 & --  & --   & $80.5$     & $85.8$      & $83.0$     & $81.5$            & $84.7$  & $83.1$    & $19.5$          \\ 
    TPSNet \cite{wang2022tpsnet}  & Reg  & $\checkmark$         & --                 & --      & --     & $86.8$      & $89.5$       & $88.1$        & $85.7$             & $87.7$      & $86.4$      & $17.9$         \\ 
    CT-Net  \cite{shao2023ct}  & Reg   & --         & $80.4$                 & $89.8$      & $84.8$     & $83.6$      & $89.2$       & $86.3$       & $82.7$     & $87.9$ 
    & $85.2$  & $13.6$           \\ 
    DPText-DETR  \cite{ye2023dptext}   & Reg   & $\checkmark$         & --                 & --      & --     & $86.4$      & $91.8$       & $89.0$         & $86.2$     & $\textbf{91.7}$ 
    & $88.8$  & $14.8$  \\
    Box2Poly  \cite{chen2024box2poly}   & Reg  & $\checkmark$         & --                 & --      & --     & $86.6$      & $90.2$       & $88.4$         & $87.5$     & $88.8$ 
    & $88.1$  & --  \\
    LRANet \cite{su2024lranet}   & Reg  & $\checkmark$     & $86.3$                 & $92.3$      & $89.2$     & $87.8$      &$90.3$       & $89.0$      & $85.5$     & $89.4$        & $87.4$   & $\textbf{37.2}$
    \\ \midrule
    TextSnake  \cite{long2018textsnake}  & CC    & $\checkmark$          & $73.9$                & $83.2$      & $78.3$     & $74.5$      & $82.7$  & $78.4$   & $85.3$  & $67.9$ 
    & $75.6$  & --           \\ 
    TextDragon  \cite{feng2019textdragon}  & CC   & $\checkmark$           & --                & --      & --     & $75.7$      & $85.6$  & $80.3$   & $82.8$  & $84.5$ 
    & $83.6$  & --           \\ 
    DRRG  \cite{2020Deep}   & CC     & $\checkmark$       & $82.3$                 & $88.1$      & $85.1$     & $84.9$      & $86.6$   & $85.8$  & $83.0$ & $86.0$ 
    & $84.5$  & $2.0$     \\ 

    ReLaText \cite{ma2021relatext}   & CC    & $\checkmark$       & $83.2$                 & $90.5$      & $86.7$     & $83.1$      & $84.8$   & $84.0$  & $83.3$ & $86.2$ 
    & $84.8$  & --     \\ 
        \midrule
      \textbf{ERRNet} & CC & -- &  86.6  & 88.2 & 87.4  & 86.1 & 90.1  & 88.1 & 85.5 & 88.9 & 87.2 & 31.7   \\
      \textbf{ERRNet} & CC & $\checkmark$  & \textbf{87.1}
      & \textbf{93.8} & \textbf{90.3}  & 87.3 & \textbf{92.6}  & 89.9 & \textbf{87.9} & 91.0 & \textbf{89.4} & 31.5   \\
      \bottomrule
    \end{tabular}
  }
  \label{tab:compare_mtc}
\end{table*}

\begin{table}[!t]
    \caption{Performance comparison on ArT. 
    $\dagger$ means the results are from the official website \cite{chng2019icdar2019}.
    }
    \centering
     \setlength\tabcolsep{5pt}
    \begin{tabular}{@{}lccc@{}}
        \toprule
        Method     & R    & P    & F   \\ \midrule
        PCR \cite{dai2021progressive}  & 66.1 & 84.0 & 74.0 \\ 
        TPSNet \cite{wang2022tpsnet}  & 73.3 & 84.3 & 78.4 \\ 
         DPText-DETR \cite{ye2023dptext}   & 73.7 & 83.0 & 78.1 \\
        LRANet \cite{su2024lranet} $^{\dagger}$ & 74.5 & 84.0 & 79.0 \\ \midrule
        \textbf{ERRNet}  & \textbf{75.5} & \textbf{84.1} & \textbf{79.6} \\
        \bottomrule
    \end{tabular}
    \label{tab:art}
\end{table}

\subsection{Datasets}
\textbf{Total-Text} \cite{ch2017total} includes horizontal, curved, and multi-oriented texts. The dataset contains $1255$ training images and $300$ test images. 
\\ \textbf{CTW1500} \cite{liu2019curved} is a challenging dataset for long curved text, which consists of $1000$ training images and $500$ test images. All text instances are annotated  by $14$-polygon.
\\ \textbf{ArT} \cite{chng2019icdar2019} 
is a large-scale multi-lingual arbitrary-shaped text detection dataset, which
includes $5603$ training images and $4563$ test images. 
\\ \textbf{MSRA-TD500} \cite{yao2012detecting} is a multi-language dataset. It consists of $300$ training images and $200$ test images.
\\ \textbf{Synth150K} \cite{liu2020abcnet} contains  $150k$ synthetic images, including about one-third of curved texts and two-thirds of multi-oriented texts.

\subsection{Implementation Details}

When training from scratch, we adopt AdamW with $1 \times 10^{-4}$ weight decay as the optimizer, and set $16$ as batch size, with $500$ training epochs for all datasets. For more comprehensive comparisons, we also pre-train our model on a mixture of SynthText-150K, MLT \cite{nayef2017icdar2017} and Total-Text for a total of $5$ epochs, and then fine-turn $300$ epochs for all datasets. The value of channel $c_v$ is $64$. For the ERR decoder, the number of layers is $3$, the maximum text instance number $n$ is $100$, and the component sequence length $t$ is $6$. For the position-supervised loss, the parameters $\alpha$ and $\gamma$ are set to $0.25$ and $2$, respectively. For data augmentation, we apply \emph{RandomCrop}, \emph{RandomRotate} and \emph{ColoJitter} to input images. In the testing stage, we set a suitable height for each dataset while keeping the original aspect ratio. The evaluation metric for the F-measure is IOU@0.5, following \cite{ye2023dptext,chen2024box2poly}. All experiments are conducted on $4$ NVIDIA RTX3090 GPUs. 

\begin{figure}[t]
\centering
    \includegraphics[width=\linewidth]{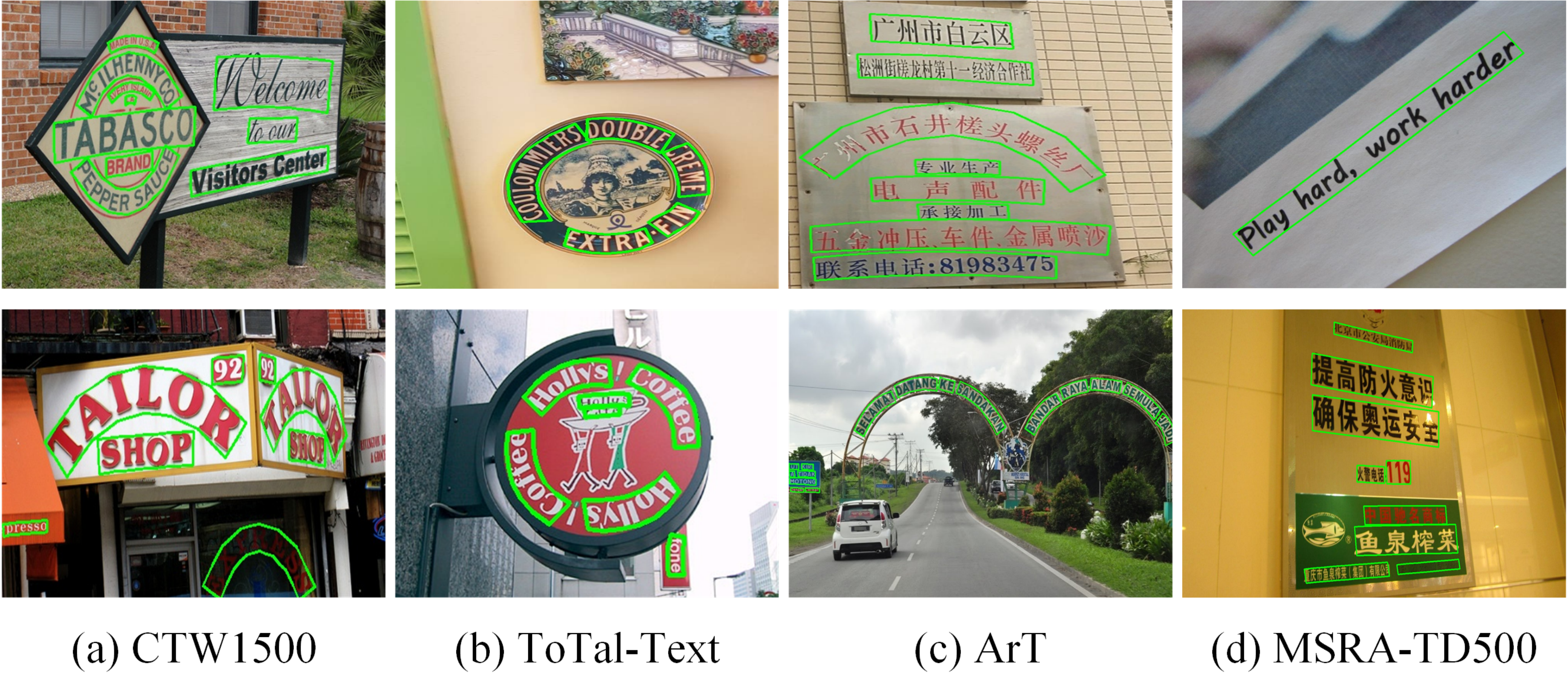}
\caption{ Qualitative detection results of our ERRNet on datasets CTW1500, Total-Text, ArT, and MSRA-TD500.}
\label{fig:visual}
\end{figure}

\subsection{Comparison with State-of-the-art Methods}
We compare ERRNet with previous methods on four challenging benchmarks. As shown in Table~\ref{tab:compare_mtc} and \ref{tab:art}, ERRNet consistently performs top-tier across the datasets. Compared to segmentation-based methods, ERRNet achieves highly competitive results even when trained from scratch. In particular, on the long curve dataset CTW1500, ERRNet outperforms the recent SOTA segmentation-based method TextBPN++ \cite{zhang2023arbitrary} even without pre-training (87.2\% vs. 86.5\% in terms of F-measure) and achieves $2.3$× faster inference speed. Meanwhile, on the other two datasets in Table~\ref{tab:compare_mtc}, the two methods also perform on par. This demonstrates the advantage of the component-level text shape representations over the pixel-level representations.

Compared to regression-based methods, ERRNet also achieves the SOTA accuracy and runs quite efficiently. Specifically, ERRNet outperforms DPText-DETR \cite{ye2023dptext} by $0.6\%$ in terms of F-measure and achieves $2.1$× faster inference speed on CTW1500. Although ERRNet is a little slower than LRANet \cite{su2024lranet}, it outperforms LRANet by margins of 1.1\%, 0.9\% and 2.0\% in terms of F-measure on MSRA-TD500, Total-Text, and CTW1500, respectively. This is because LRANet has difficulty in accurately capturing the diverse geometric layouts of the text through a single perception. Moreover, on even large ArT dataset, our ERRNet outperforms DPText-DETR and LRANet by 1.5\% and 0.6\% in terms of F-measure respectively, again demonstrating the superiority of ERRNet.

In comparison with the previous CC-based methods, ERRNet dramatically surpasses them in both accuracy and speed. Specifically, ERRNet surpasses DRRG \cite{2020Deep} by $4.9\%$ in terms of F-measure on CTW1500 and achieves $16$× inference acceleration. This is mainly attributed to our explicit relational reasoning, which not only reduces the difficulty of component content learning but also is post-processing-free. Some detection visualizations are shown in Fig.~\ref{fig:visual}. ERRNet performs well on long, small, and curved text instances.

\begin{table}[t]
    \caption{Performance gains of our ERR decoder.}
    \centering
     \setlength\tabcolsep{10pt}
    \begin{tabular}{@{}lcccc@{}}
        \toprule
        Dataset & ERR Decoder & R    & P  & F \\ \midrule
        CTW1500 & -- & 82.1 & 87.5 & 84.7 \\ 
        CTW1500 & \checkmark  & \textbf{85.5} & \textbf{88.9} & \textbf{87.2} \\ \midrule
        Total-Text & -- & 83.8 & 88.5 & 86.1 \\ 
        Total-Text & \checkmark  & \textbf{86.1} & \textbf{90.1} & \textbf{88.1} \\ 
        \bottomrule
    \end{tabular}
    \label{tab:decoder}
\end{table}

\begin{table}[t]
    \caption{Ablation study of our position-supervised classification loss (PSC in short) on Total-Text. $\ddagger$ means the results obtained after reproducing the model.}
    \centering
     \setlength\tabcolsep{4pt}
    \begin{tabular}{@{}lcccc@{}}
        \toprule
        Method & PSC & R    & P    & F   \\ \midrule
        DPText-DETR \cite{ye2023dptext} $^{\ddagger}$  & -- & 85.1 & 88.4 & 86.7 \\ 
        DPText-DETR \cite{ye2023dptext}  $^{\dagger}$ & \checkmark & \textbf{85.7} & \textbf{89.1} & \textbf{87.4} \\ \midrule
        ERRNet (Ours) & -- & 85.1 & 89.5 & 87.3 \\ 
        ERRNet (Ours) & \checkmark & \textbf{86.1} & \textbf{90.1} & \textbf{88.1} \\ 
        \bottomrule
    \end{tabular}
    \label{tab:psc_loss}
\end{table}

\begin{table}[t]
    \caption{Experimental results for different reasoning mothods. ERR and IRR denote explicit relational reasoning and implicit relational reasoning, respectively.}
    \centering
     \setlength\tabcolsep{10pt}
    \begin{tabular}{@{}lcccc@{}}
        \toprule
        Dataset & Method & R    & P    & F   \\ \midrule
        CTW1500 & IRR & 84.9 & 87.7 & 86.3 \\ 
        CTW1500 & ERR & \textbf{85.5} & \textbf{88.9} & \textbf{87.2} \\ \midrule
        Total-Text & IRR & 85.3 & 88.8 & 87.0 \\ 
        Total-Text & ERR & \textbf{86.1} & \textbf{90.1} & \textbf{88.1} \\ 
        \bottomrule
    \end{tabular}
    \label{tab:track_vs_graph}
\end{table}

\begin{table}[t]
    \caption{Performance of ERRNet with different input sizes.}
    \centering
     \setlength\tabcolsep{8pt}
    \begin{tabular}{@{}lccccc@{}}
        \toprule
        Dataset & Input & R    & P    & F & FPS  \\ \midrule
        CTW1500 & 512 & 85.3 & 88.1 & 86.7 & \textbf{41.7} \\ 
        CTW1500 & 608 & 85.4 & 88.5 & 86.9 & 36.6 \\ 
        CTW1500 & 704 & \textbf{85.5} & \textbf{88.9} & \textbf{87.2} & 31.5 \\       
        \midrule
        Total-Text & 608 & 83.5 & 86.9 & 85.2 & \textbf{34.8} \\ 
        Total-Text & 800 & 85.9 & 89.3 & 87.6 & 28.9 \\ 
        Total-Text & 1000 & \textbf{86.1} & \textbf{90.1} & \textbf{88.1} & 21.7 \\ 
        \bottomrule
    \end{tabular}
    \label{tab:sizes}
\end{table}

\subsection{Ablation Study}

We perform ablation studies on CTW1500 and Total-Text datasets, without pre-training applied by default. 

\noindent \textbf{ERR Decoder.}
We conduct experiments to verify the influence of our ERR decoder. 
The results are listed in Table~\ref{tab:decoder}. The ERR decoder achieves improvements of $2.5\%$ and $2.0\%$ in F-measure on CTW1500 and Total-Text, respectively. Remarkably, the improvements are mainly contributed by recall ($3.4\%$ on CTW1500 and $3.7\%$ on Total-Text), mainly because the decoder efficiently refines the cluttered components with low confidence from the TCI module. Moreover, the performance remains competitive even without the decoder, indicating the effectiveness of our text component initialization module.

\noindent \textbf{Position-Supervised Classification Loss.}
We ablate our position-supervised classification loss on Total-Text to assess its impact. As shown in Table~\ref{tab:psc_loss}, the position-supervised loss improves the F-measure of ERRNet by $0.8\%$. This demonstrates the effectiveness of dynamically adjusting classification targets based on the prediction quality. To verify the generality of our PAC Loss, we also apply it to DPText-DETR \cite{ye2023dptext}. It can be seen that embedding this loss improves DPText-DETR by $0.6\%$, $0.7\%$, and $0.7\%$ in the precision, recall, and F-measure, respectively. 

\noindent \textbf{Explicit Relational Reasoning.}
We adopt explicit relational reasoning to pre-define the component relationships in each frame. To explore the effectiveness of explicit relational reasoning, we design a variant of implicit relational reasoning, i.e., predicting component associative relationships via a prediction head. Indeed, this variant adopts the perspective of scene graph generation in ReLaText \cite{ma2021relatext} to implicitly model component relationships. As shown in Table~\ref{tab:track_vs_graph}, explicit relational reasoning consistently outperforms the implicit relational reasoning, mainly because pre-defined relationships introduce more prior information, which reduces the difficulty of learning component relationships.

\noindent \textbf{Different Input Image Sizes.}
To assess the influence of image size and  making a proper trade-off between accuracy and speed, we evaluate ERRNet with different short side lengths. As shown in Table~\ref{tab:sizes}, ERRNet is robust to changes in image size, with the F-measure fluctuating by only $0.3\%$ on CTW1500 when the size changes from 704 to 608, and by $0.5\%$ on Total-Text when the size changes from 1000 to 800. 

\section{Conclusion}

In this paper, we have presented ERRNet, an accurate and efficient CC-based text detector. For the first time, ERRNet groups text components from a tracking perspective, explicitly defining the relationships between components along both the spatial and temporal dimensions. Therefore, the detection task is elegantly transformed into a tracking task, and post-processing-free prediction is achieved. Additionally, a position-supervised loss is introduced to guide ERRNet towards more consistent task-aligned learning. Experiments conducted on public benchmarks have confirmed the effectiveness of the proposed ERRNet, which shows leading accuracy and top-ranked inference speed. Given its effectiveness and efficiency, we are interested in extending our approach of explicit relational modeling to the scene text understanding task \cite{liang2024layoutformer}, i.e., explicitly modeling the relationships between words, sentences, and paragraphs for post-processing-free prediction.

\section*{Acknowledgements}

This work was supported by National Natural Science Foundation of China (62372170, 62172103)

\appendix

\section{More Experiments}

\noindent \textbf{Component sequence length.}
We explore the influence of component sequence length, denoted as $t$ (i.e., the number of frames), on model performance.
As $t$ increases from $4$ to $6$, the F-measure improves by $0.4\%$, mainly because  short components cannot fit the complex text well. As $t$ continues to increase, there is no further performance improvement. Therefore, we set $t=6$ to balance computational complexity and representation quality.

\begin{table}[h]
        \vspace{-1mm}   
    \caption{Performance of ERRNet with different number of component sequence length on CTW1500.}
        \vspace{-2mm}   
    \centering
     \setlength\tabcolsep{22pt}
    \begin{tabular}{@{}lccc@{}}
        \toprule
        $t$   & R    & P    & F  \\ \midrule
        4 & 85.0 & 88.7 & 86.8 \\ 
        6 & 85.5 & \textbf{88.9} & \textbf{87.2} \\ 
        8  & \textbf{86.4} & 88.0 & \textbf{87.2} \\ 
        \bottomrule
    \end{tabular}
    \label{tab:bezier_vs_bspline}
        \vspace{-1mm}  
\end{table}

\begin{figure}[t]
    \centering
    \subfigure [Bezier]
    {
        \includegraphics[width=0.22\textwidth,height=2.0cm]{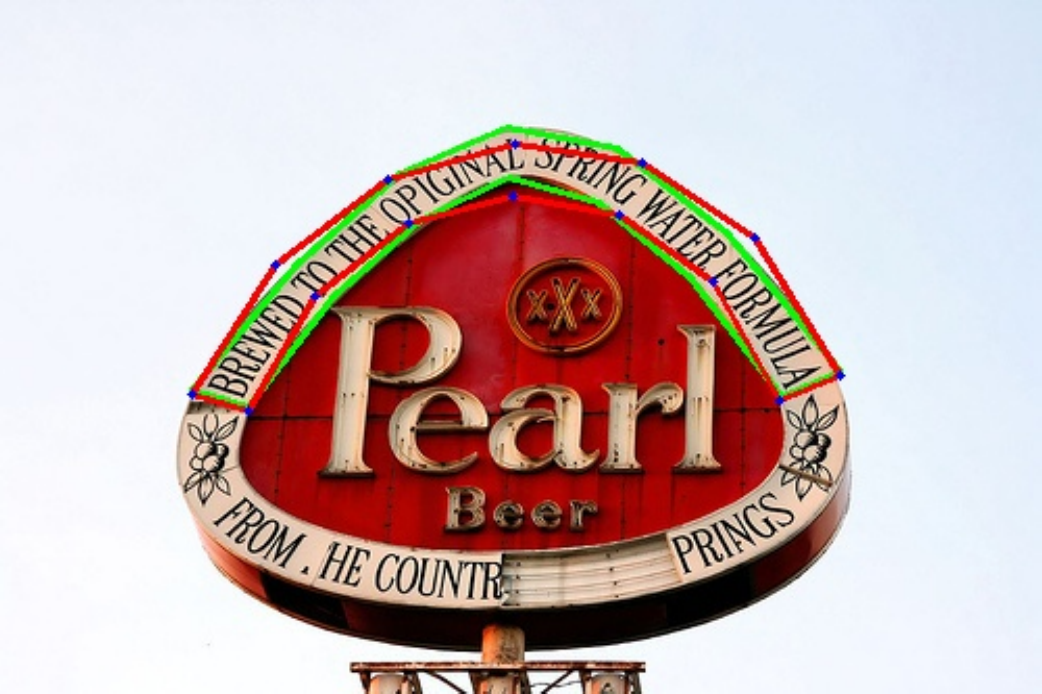}
    }
    \subfigure [B-spline]
    {
        \includegraphics[width=0.22\textwidth,height=2.0cm]{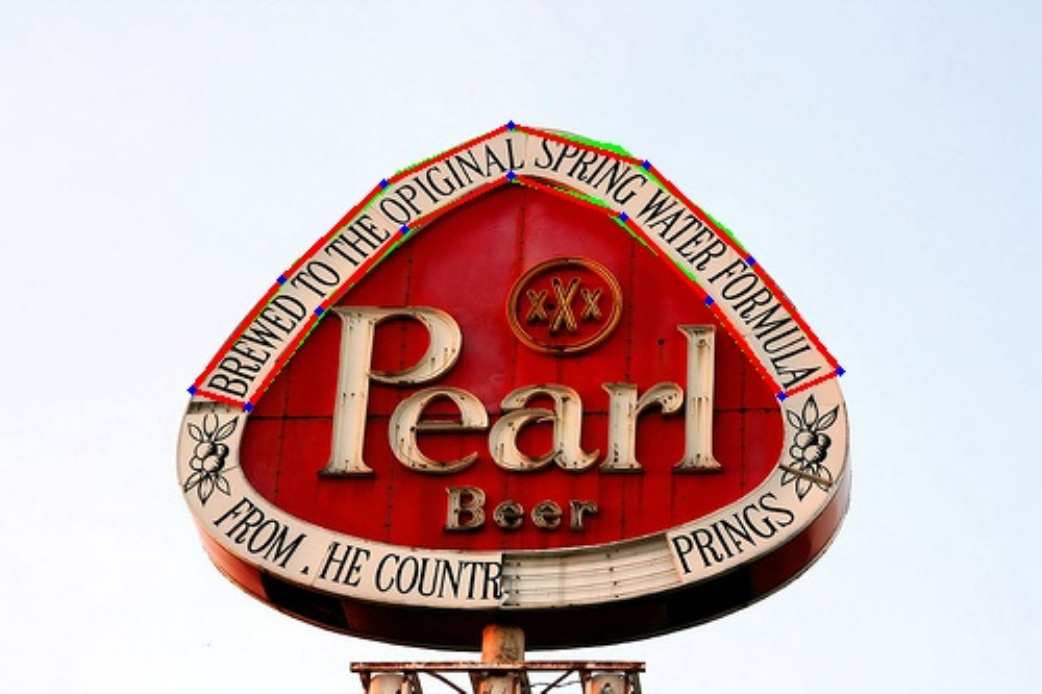}
    } 
    \vspace{-3mm}   
    \caption{Visualization comparison of
    different interpolation methods. The green polygons are ground-truth and the red lines are interpolated curves.}
    \label{fig:bezier_vs_bspline}
    \vspace{-3mm}   
\end{figure}

\noindent  \textbf{B-spline interpolation vs. Bezier interpolation.}
In the TCI module, we utilize B-spline interpolation to resample text instance vertices at uniform intervals. Compared with Bezier interpolation \cite{liu2020abcnet},  which is currently the most widely used method for interpolating text contours, B-spline interpolation offers superior fitting of complex text shapes due to its robust local control capability, as depicted in Fig.~\ref{fig:bezier_vs_bspline}. For quantitative comparison, we select a complex text subset (total 114 samples) following the method in \cite{zhu2021fourier}.
As shown in Table~\ref{tab:bezier_vs_bspline}, where PIoU (Polygon intersection over union ) measures the overlap between reconstructed and ground-truth text regions, B-spline interpolation demonstrates superior fitting ability for complex text shapes (99.2\% vs. 97.4\% in terms of PIoU), resulting in a 1.1\% improvement in F-measure.

\begin{table}[h]
        \vspace{-1mm}   
    \caption{Comparison of Bezier interpolation and B-spline interpolation on a highly curved text subset of CTW1500.}
            \vspace{-2mm}   
    \centering
     \setlength\tabcolsep{14pt}
    \begin{tabular}{@{}lcccc@{}}
        \toprule
        Reps   & R    & P    & F  & PIoU \\ \midrule
        Bezier & 82.1 & 84.3 & 83.2 & 97.4 \\ 
        B-spline  & \textbf{82.3} & \textbf{86.4} & \textbf{84.3} & \textbf{99.2} \\ 
        \bottomrule
    \end{tabular}
    \label{tab:bezier_vs_bspline}
        \vspace{-1mm}  
\end{table}

\begin{table}[h]
        \vspace{-1mm}   
    \caption{Performance comparison of different metrics to measure  localization quality. BIoU refers to the intersection over union of the minimum bounding rectangle.}
        \vspace{-2mm}   
    \centering
     \setlength\tabcolsep{22pt}
    \begin{tabular}{@{}lccc@{}}
        \toprule
        Types   & R    & P    & F   \\ \midrule
        BIoU & 84.9 & 87.8 & 86.4  \\ 
        PIoU  & \textbf{85.5} & \textbf{88.9} & \textbf{87.2}  \\ 
        \bottomrule
    \end{tabular}
    \label{tab:ablation_piou}
            \vspace{-3mm}   
\end{table}

\noindent  \textbf{Polygon Monte-Carlo Method.}
To investigate the effectiveness of our  Polygon Monte-Carlo method, we compare it with the minimum bounding rectangle evaluation method. Both allow quick evaluation of the location quality through vectorized computations. As shown in Table~\ref{tab:ablation_piou}, our method exhibits superior performance, indicating that it more accurately measures localization quality.

\section{More Visualizations}

\begin{figure*}[!t]
\centering
\includegraphics[width=\textwidth]{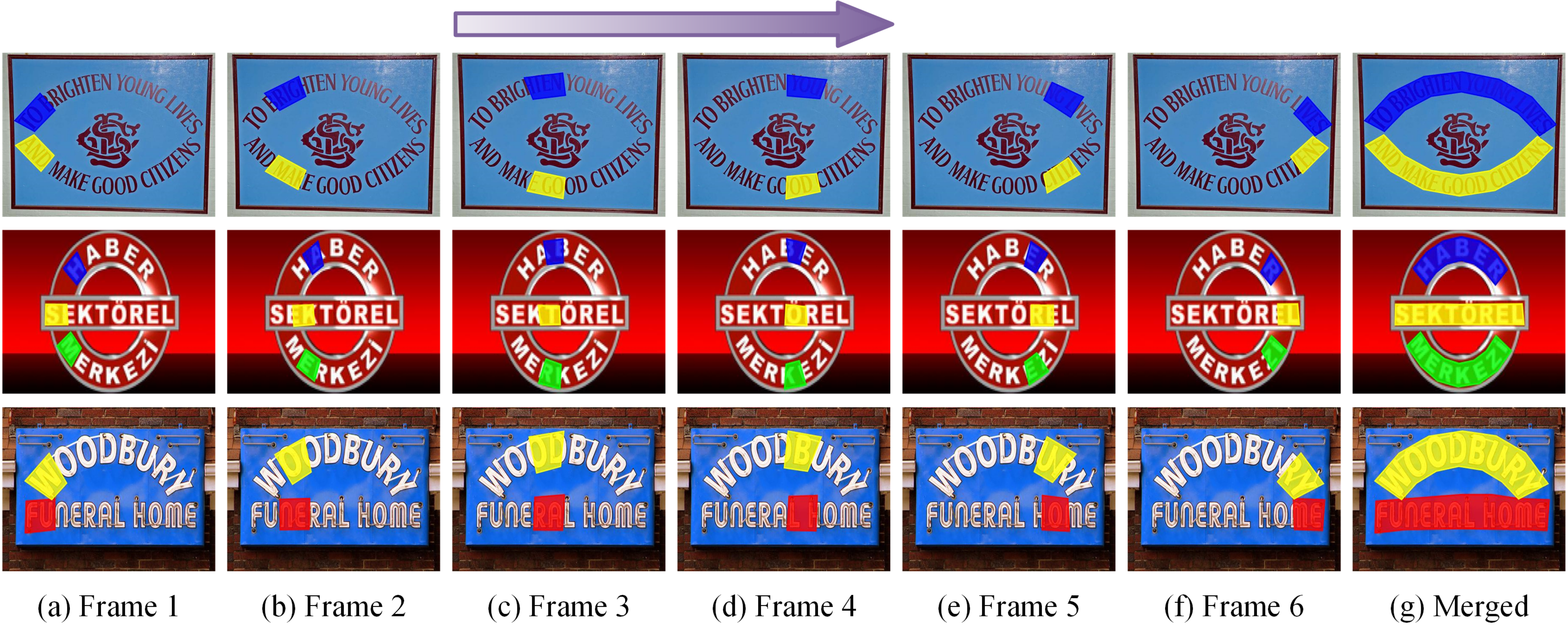}
\caption{Visualization of our ERRNet, which models each text instance as multiple text components in sequential movement.
}
\label{fig:tracking}
\end{figure*}

\noindent  \textbf{Visualization of Tracking Perspective.}
We model text shapes from a tracking perspective, where each text instance can be viewed as multiple text components in sequential movement. The visualization results in Fig.~\ref{fig:tracking} validate the feasibility of this perspective. 

\begin{figure*}[!t]
\centering
\includegraphics[width=\textwidth]{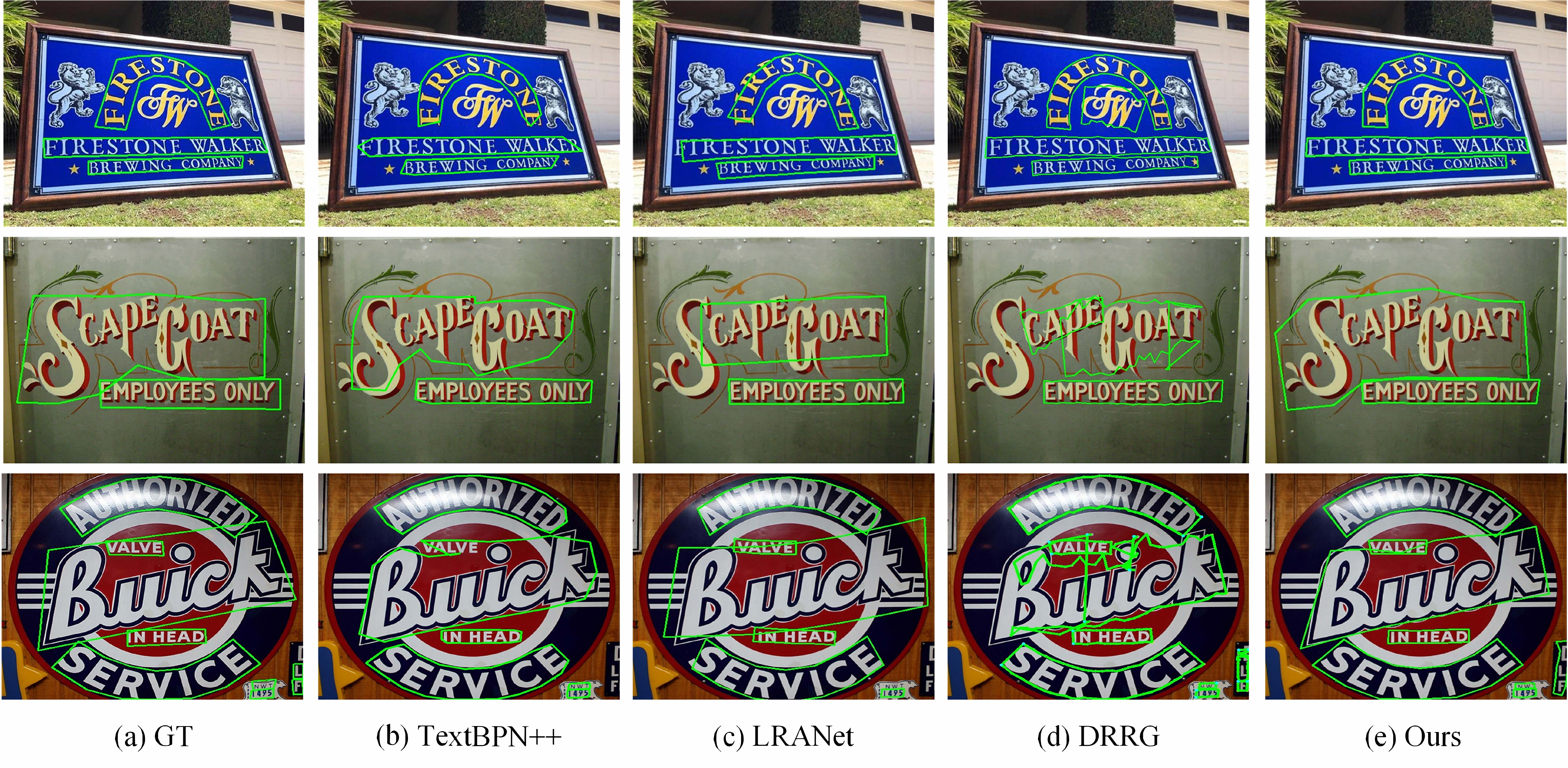}
\caption{More visualization comparison with the current SOTA across three different types: (a) Segmentation-based method (TextBPN++), (b) Regression-based method (LRANet), and (c) Connected component-based method (DRRG). 
}
\label{fig:m_compare}
\end{figure*}

\noindent  \textbf{Qualitative Comparisons.}
We further qualitatively compare our ERRNet with leading methods of different types. As shown in Fig.~\ref{fig:m_compare}, compared to the segmentation-based method TextBPN++ \cite{zhang2023arbitrary}, our method can accurately detect overlapping text. Compared to the regression-based method LRANet \cite{su2024lranet}, ERRNet shows better adaptability to the variability of text sizes and shapes. Furthermore, compared to the CC-based method DRRG \cite{2020Deep}, our method excels in predicting the locations of components and the relationships between them. Notably, some predictions made by our method are even more accurate than the ground-truth.


\newpage

\bibliography{aaai25}

\end{document}